%% file: main.tex
\documentclass[10pt,twocolumn,letterpaper]{article}

\usepackage{wacv}              

\input{preamble}

\definecolor{wacvblue}{rgb}{0.21,0.49,0.74}

\def\wacvPaperID{2458} 
\def\confName{WACV}
\def\confYear{2026}

\title{CountingDINO \includegraphics[height=0.8em]{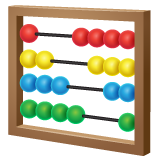}\includegraphics[height=1em]{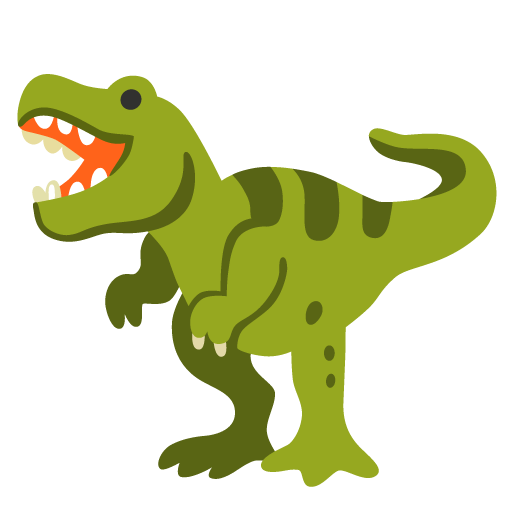}\\A Training-free Pipeline for Class-Agnostic Counting using Unsupervised Backbones}

\author{Giacomo Pacini$^{*1,2}$,
Lorenzo Bianchi$^{*1,2}$,
Luca Ciampi$^1$, 
Nicola Messina$^1$, \\
Giuseppe Amato$^1$, 
Fabrizio Falchi$^1$ \\
\quad$^1$ISTI-CNR, Italy\quad$^2$University of Pisa, Italy\quad$^*$Equal contribution\\
{\tt\small \{name.surname\}@isti.cnr.it}}

\begin{document}

\maketitle

\input{sec/0_abstract}    
\input{sec/1_introduction}
\input{sec/2_related}
\input{sec/3_method}
\input{sec/4_experiments}

\input{sec/5_conclusions}
\input{sec/6_ack}

{
    \small
    \bibliographystyle{ieeenat_fullname}
    \bibliography{main}
}

\include{sec/X_suppl}

\end{document}

%% file: preamble.tex
\usepackage{ifthen}
\PassOptionsToPackage{table}{xcolor}

\newboolean{reviewVersion}
\setboolean{reviewVersion}{true}
\newboolean{shortenedVersion}
\setboolean{shortenedVersion}{true}  

\usepackage[pagebackref,breaklinks,colorlinks,allcolors=wacvblue]{hyperref}
\usepackage{adjustbox}
\usepackage{array}

\usepackage{graphicx}
\usepackage{booktabs}
\usepackage{color}
\usepackage{colortbl}

\usepackage{algorithm}
\usepackage{algpseudocode}

\usepackage{comment}
\usepackage{amssymb}
\usepackage{tabularx}
\usepackage{booktabs} 
\usepackage{pifont}

\usepackage{arydshln}
\usepackage{float}

\usepackage{orcidlink}

\usepackage[accsupp]{axessibility}  

\definecolor{LightCyan}{rgb}{0.88,0.95,1}
\newcommand{\ours}{CountingDINO\xspace}
\DeclareMathOperator*{\argmax}{arg\,max}  

\newcommand{\figimg}[1]{\includegraphics[width=\figw, height=2.5cm]{#1}}

%% file: sec/0_abstract.tex
\begin{abstract}
Class-agnostic counting (CAC) aims to estimate the number of objects in images without being restricted to predefined categories. However, while current exemplar-based CAC methods offer flexibility at inference time, they still heavily rely on labeled data for training, which limits scalability and generalization to many downstream use cases. In this paper, we introduce \ours, the first \textit{training-free} exemplar-based CAC framework that exploits a \textit{fully unsupervised} feature extractor. Specifically, our approach employs self-supervised vision-only backbones to extract object-aware features, and it eliminates the need for annotated data throughout the entire proposed pipeline.
At inference time, we extract latent object prototypes via ROI-Align from DINO features and use them as convolutional kernels to generate similarity maps. These are then transformed into density maps through a simple yet effective normalization scheme. 
We evaluate our approach on the FSC-147 and CARPK benchmarks, where we consistently outperform a baseline based on an SOTA unsupervised object detector under the same label- and training-free setting. Additionally, we achieve competitive results -- and in some cases surpass -- training-free methods that rely on supervised backbones, non-training-free unsupervised methods, as well as several fully supervised SOTA approaches.
This demonstrates that label- and training-free CAC can be both scalable and effective. Website: \url{https://lorebianchi98.github.io/CountingDINO/}.
\end{abstract}

%% file: sec/1_introduction.tex
\section{Introduction}
\label{sec:intro}
\ifthenelse{\boolean{shortenedVersion}}{
Class-agnostic counting (CAC) removes the need for class-specific retraining by allowing users to define target categories at inference~\cite{DBLP:conf/cvpr/RanjanSNH21}. This recent paradigm addresses the inherent limitations of class-specific counting approaches, which rely on predefined object types, such as vehicles~\cite{DBLP:conf/iccv/ZhangWCM17,DBLP:conf/iscc/AmatoCFG19}, people~\cite{DBLP:conf/cvpr/LiuSF19,DBLP:journals/eswa/BenedettoCCFGA22}, cells~\cite{DBLP:journals/mia/CiampiCTMLSAPG22,DBLP:conf/eccv/XueRHB16}, or animals~\cite{DBLP:conf/eccv/ArtetaLZ16,DBLP:journals/cea/TianGCWLM19}. 

However, CAC models still rely heavily on annotated training data. First, most existing approaches extract image features using backbones pre-trained on large-scale labeled datasets such as ImageNet~\cite{DBLP:conf/cvpr/HeZRS16}. Second, they require supervised training on annotated datasets, including thousands of dot annotations and bounding boxes, limiting scalability.
}
{
Class-agnostic counting (CAC) aims to count instances of arbitrary object classes beyond those encountered during training~\cite{DBLP:conf/cvpr/RanjanSNH21}. This recent paradigm addresses the inherent limitations of traditional class-specific counting approaches, which rely on models trained for predefined object types, such as vehicles~\cite{DBLP:conf/iccv/ZhangWCM17,DBLP:conf/iscc/AmatoCFG19}, people~\cite{DBLP:conf/cvpr/LiuSF19,DBLP:journals/eswa/BenedettoCCFGA22}, cells~\cite{DBLP:journals/mia/CiampiCTMLSAPG22,DBLP:conf/eccv/XueRHB16}, or animals~\cite{DBLP:conf/eccv/ArtetaLZ16,DBLP:journals/cea/TianGCWLM19}. Unlike these methods, CAC allows users to dynamically define target categories during inference, removing the need to retrain deep learning-based networks with class-specific annotated datasets.

However, CAC still relies heavily on human annotations. First, most existing approaches extract image features using backbones pre-trained on large-scale labeled datasets such as ImageNet~\cite{DBLP:conf/cvpr/HeZRS16}. Second, they require supervised training on annotated datasets containing thousands of objects across hundreds of categories. 
In particular, the most widely adopted CAC paradigm -- the exemplar-based approach -- typically involves two types of supervision: dot annotations and a small number of bounding boxes. 
}
\begin{figure}[t]
    \centering
    \includegraphics[width=\linewidth]{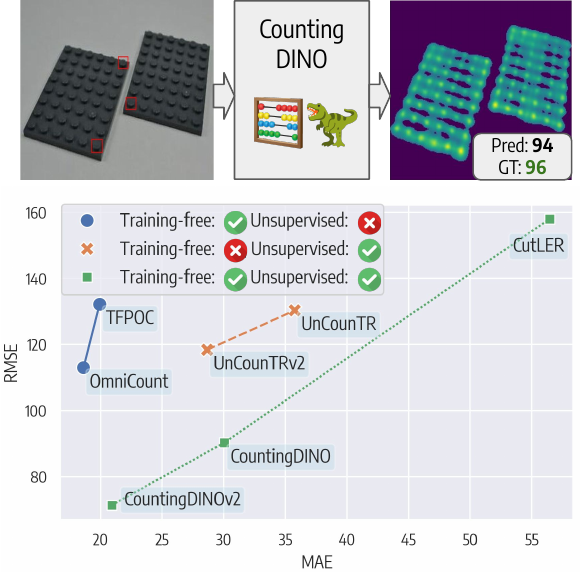}
    \caption{CountingDINO is a training-free and unsupervised class-agnostic counting method. The instance running with DINOv2 is able to outperform a training-free unsupervised detection-based baseline (CutLER), on both MAE and RMSE metrics (lower is better). Notably, it reaches comparable results (and even outperforms) previous non-training-free or supervised methods.}
    \label{fig:method}
\end{figure}
Dot annotations, placed on object centroids, are used to generate density maps that serve as training targets for a regression network learning to map image features to object density estimates~\cite{DBLP:conf/nips/LempitskyZ10}. Bounding boxes, on the other hand, are used to localize a small set of visual prototypes, known as \textit{exemplars}, which represent the object category to be counted~\cite{DBLP:conf/cvpr/RanjanSNH21}.
Particularly, the requirement for thousands of dot annotations poses a major obstacle to creating large-scale datasets. As a result, only a few datasets are available, and even these often exhibit notable limitations~\cite{10944113}.
Many previous works tried to approach this level of \textit{un}supervision for the counting task. Nevertheless, in the end, they surrendered to performing supervised finetuning employing a variable amount of ground-truth density maps \cite{Liu_2022_BMVC_countr}, or employing upstream vision backbones or segmentation models requiring costly annotations to be trained (e.g., SAM) \cite{DBLP:conf/wacv/Shi0Z24}.

In this paper, we introduce \ours, the first \textit{training-free} exemplar-based CAC framework that eliminates the need for labeled data at any stage. \ours shows that it is possible to effectively employ modern, powerful self-supervised vision features without any further downstream training or finetuning, or without relying on any kind of manually-labeled data. \ours builds upon a smart and simple design, and shows that smart inference techniques can enable off-the-shelf self-supervised vision backbones to effectively count arbitrary objects in an image.
Our image feature extraction backbone builds on DINO~\cite{caron2021emerging,oquab2023dinov2,darcet2024vision}, which is trained in a self-supervised fashion.
At inference time, we extract exemplar features by applying ROI-Align~\cite{8372616} to user-provided bounding boxes using the same DINO-computed image feature maps. These exemplar features are then used as depthwise convolutional kernels over the image features, generating similarity maps that highlight regions matching the exemplars. 
We average these similarity maps across exemplars to produce a global, informative response, which is then converted into a density map through a simple yet effective normalization scheme. Final object counts are computed by integrating the density map. 
Furthermore, to address the limitations of the spatial resolution of DINO with small objects, we partition the image into non-overlapping quadrants, apply feature extraction independently, and aggregate the resulting maps. 
We validate our method on the gold-standard CAC benchmarks FSC-147~\cite{DBLP:conf/cvpr/RanjanSNH21} and CARPK~\cite{hsieh2017drone}, consistently outperforming a baseline based on an SOTA unsupervised object detector under the same label- and training-free setting. We also compare it to training-free methods relying on supervised backbones, non-training-free unsupervised methods, and fully supervised approaches, demonstrating that our method remains competitive despite not using supervision or training at any stage.

To summarize, we propose the following contributions:

\begin{itemize}
    \item We propose \ours, the first \textit{training-free} class-agnostic counting framework that uses self-supervised backbones to extract object-aware features, removing any need for annotated data or any form of training.
    \item 
    We assess our approach on the FSC-147 and CARPK benchmarks. \ours outperforms a baseline under the same label- and training-free setting and achieves results that are comparable to -- and in some cases surpass  -- recent training-free methods with supervised backbones, unsupervised approaches (that are not training-free), as well as fully supervised approaches.
    \item We conduct ablation studies to assess the contribution of each core component in our model.
\end{itemize}

%% file: sec/2_related.tex
\section{Related Works}
\label{sec:related}


\subsection{Exemplar-based Class-agnostic Counting}
Counting category-specific objects is a longstanding task in computer vision with broad real-world applications -- e.g., counting of people~\cite{DBLP:conf/cvpr/LiuSF19,DBLP:journals/eswa/BenedettoCCFGA22}, vehicles~\cite{DBLP:conf/iccv/ZhangWCM17,DBLP:conf/iscc/AmatoCFG19}, insects~\cite{DBLP:journals/ecoi/CiampiZICBFAC23,DBLP:journals/cea/Bereciartua-Perez22}, or biological structures~\cite{DBLP:journals/mia/CiampiCTMLSAPG22,DBLP:conf/eccv/XueRHB16}.
Among existing counting approaches, density map regression -- estimating counts via feature-to-density mapping -- has proven effective in crowded scenes~\cite{DBLP:conf/eccv/ArtetaLZ16,DBLP:conf/nips/LempitskyZ10}, outperforming detection-based techniques~\cite{DBLP:journals/cea/TianGCWLM19,DBLP:conf/iscc/AmatoCFG19}. However, their reliance on category-specific annotations and training limits scalability and generalization.

Class-agnostic counting (CAC) has recently shifted the object counting paradigm toward open-world contexts, enabling models to handle arbitrary categories unseen during training. In this setting, users are no longer restricted to predefined categories and can instead specify novel object classes at inference time by providing visual exemplars -- typically in the form of three bounding boxes enclosing visual prototypes within the same input image~\cite{DBLP:conf/cvpr/RanjanSNH21}. The seminal FamNet~\cite{DBLP:conf/cvpr/RanjanSNH21} combines multi-scale feature extraction with exemplar-image correlation via inner product operations for density map prediction. RCAC~\cite{DBLP:conf/eccv/GongZ0DS22} follows a similar architecture but introduces a feature augmentation module that synthesizes exemplars with varied colors, shapes, and scales to enhance diversity.
BMNet~\cite{DBLP:conf/cvpr/Shi0FL022} highlights the limitations of inner product-based correlation and instead proposes a learnable bilinear similarity loss, inspired by metric learning, for better matching.
CounTR~\cite{Liu_2022_BMVC_countr} proposes a method for counting elements in images of arbitrary semantic categories, by employing a two-stage training pipeline: a MAE-based self-supervised stage, which learns powerful visual features, and a fully-supervised final stage. 
LOCA~\cite{DBLP:conf/iccv/EukicLZK23} introduces an object prototype extraction module that iteratively refines exemplar features via cross-attention, while PseCo~\cite{DBLP:conf/cvpr/HuangD0ZS24} relies on the popular SAM for instance segmentation~\cite{DBLP:conf/iccv/KirillovMRMRGXW23}. 
CACViT~\cite{DBLP:conf/aaai/WangX0024} leverages a single pre-trained Vision Transformer, using its attention mechanism for both feature extraction and matching.
DAVE~\cite{DBLP:conf/cvpr/PelhanLZK24} proposes a two-stage detect-and-verify framework: first identifying candidate boxes via density maps and then verifying them using exemplar-based clustering. 
In contrast to these supervised methods, UnCounTR~\cite{knobel2024uncountr} is trained without manual annotations by using self-collages generated from unsupervised data and DINO features. Instead, TFPOC~\cite{DBLP:conf/wacv/Shi0Z24} and OmniCount~\cite{DBLP:journals/corr/abs-2403-05435} propose training-free pipelines using SAM without point-level supervision — though SAM is label-trained, making the approach not fully label-free.
A recent survey of CAC methods is available in~\cite{DBLP:journals/corr/abs-2501-19184}. 

\subsection{Unsupervised Vision Backbones}
Recent advances in unsupervised, especially self-supervised, learning have produced models with highly effective representations for classification and localization. The DINO family~\cite{caron2021emerging,oquab2023dinov2,darcet2024vision}, based on vision transformers~\cite{dosovitskiy2021an} and trained via self-distillation, demonstrates that patch-level features learned without labels can capture rich semantic information, outperforming CLIP-based models, which rely on weak supervision from image-text pairs. Alongside DINO, masked autoencoders (MAE)~\cite{he2022masked} offer a self-supervised approach by training transformers to reconstruct masked patches, fostering global visual understanding.

In this work, we propose the first \textit{training-free} exemplar-based CAC framework that leverages \textit{fully unsupervised} vision backbones to extract features from both images and exemplars. Our pipeline requires no labeled data or supervision at any stage, neither for training the image feature extractor nor for generating the density maps.

%% file: sec/3_method.tex
\section{Method}
\label{sec:method}
\begin{figure*}[t]
    \centering
    \includegraphics[width=\linewidth]{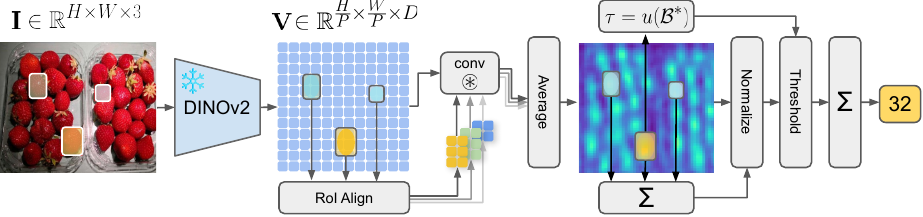}
    \caption{\textbf{Overview of \ours}. Given an image $I$ and $N$ exemplar boxes, we extract features using the DINO-based visual backbone and apply each exemplar as a convolutional kernel over the image feature map to obtain similarity maps. These are aggregated, normalized into a density map using spatial priors, and thresholded before integration to produce the final count.}
    \label{fig:method}
\end{figure*}

Given an image 
and a small set of exemplar bounding boxes from the same image -- each containing a visual prototype of the object we want to count -- we aim to estimate the total number of instances of that object class present in the image,  without relying on human-labeled data at any stage.
We claim that, by leveraging the object understanding capabilities of DINO~\cite{caron2021emerging,oquab2023dinov2,darcet2024vision} -- a family of self-supervised vision-only backbones exhibiting strong localized understanding capabilities -- we can build a simple yet effective exemplar-based CAC model. While the core of the proposed approach is mainly constituted by the powerful DINO features, the proposed pipeline integrates simple yet clever downstream processing steps that can count objects given a few exemplars. We report a detailed schema in Fig.~\ref{fig:method}, and we describe each component in the following sections. 


\subsection{DINO-based Feature Extraction}
\label{ssec:feature_extraction}
Let $\mathbf{I} \in \mathbb{R}^{H \times W \times 3}$ be an input image, and let $\psi_\text{v}$ denote a DINO-based self-supervised visual backbone. Passing the image through $\psi_\text{v}$ yields a dense feature map:
$
\mathbf{V} = \psi_\text{v}(\mathbf{I}) \in \mathbb{R}^{L \times V \times D},
$
where $L=\frac{H}{P}$ and $V=\frac{W}{P}$ represent the spatial dimensions of the feature map, with $P$ being the patch size and $D$ the dimensionality of the feature embeddings. Each location in $\mathbf{V}$ encodes a feature vector corresponding to a specific image patch.

We also extract the representation from all the available exemplars. Each exemplar is specified by a bounding box in the original image space, defined by its top-left and bottom-right coordinates as $\mathbf{b} = (x_1, y_1, x_2, y_2)$, where $x_1, y_1, x_2, y_2 \in \mathbb{R}$. 
We simply perform ROI-Align on $\mathbf{V}$ using the bounding box $\mathbf{b}$, by carefully downsampling the coordinates of $\mathbf{b}$ to $\mathbf{\tilde{b}} =  (\lfloor L x_1 \rfloor, \lfloor V y_1 \rfloor, \lceil L x_2 \rceil, \lceil V y_2 \rceil)$. The output of this operation is a feature $\mathbf{R}^{\mathbf{\tilde{b}}}$ that captures the visual appearance of the exemplar $\mathbf{\tilde{b}}$ in the embedding space.


To refine the exemplar representation, we apply a soft spatial prior in the form of an elliptical weighting mask centered within the bounding box. This mask attenuates peripheral regions and emphasizes the center under the assumption that exemplar objects are typically centered within their boxes. Formally, we define a mask $\mathbf{M}^\mathbf{\tilde{b}} \in \mathbb{R}^{w(\mathbf{\tilde{b}}) \times h(\mathbf{\tilde{b}})}$, where each entry $m^\mathbf{\tilde{b}}_{ij}$ indicates the proportion of the corresponding feature cell in the bounding box $\mathbf{\tilde{b}}$ that lies within an ellipse centered in it. 
The impact of this spatial prior and the underlying assumption is further analyzed through ablation experiments in Section~\ref{ssec:ablation}.

\subsection{Similarity Map Generation}
In this stage, we create a 2D similarity map where each value reflects how closely a local image region matches the exemplar in appearance and structure. To obtain this, we use each pooled exemplar feature $\mathbf{R}^{\mathbf{\tilde{b}}}$ as a convolutional kernel on the map $\mathbf{V}$, i.e., $\mathbf{S}^\mathbf{\tilde{b}} = \text{Conv2D}(\mathbf{V}, \mathbf{R}^{\mathbf{\tilde{b}}}) \in \mathbb{R}^{\bar{L} \times \bar{V}}$, where $\bar{L}$ and $\bar{V}$ are the spatial dimensions of the output map. 

We repeat this process for each of the $N$ exemplar bounding boxes $\{\mathbf{\tilde{b}}_i\}_{i=1}^N$, resulting in $N$ similarity maps ${\mathbf{S}_1, \mathbf{S}_2, \dots, \mathbf{S}_N}$. 
Finally, we average the aligned maps to obtain a single aggregated similarity map $\mathbf{S} = \frac{1}{N}\sum_{i=1}^N \mathbf{S}_i$, which reflects the overall likelihood of object occurrence across all exemplars.

\subsection{Density Map Normalization}  
The aggregated similarity map $\textbf{S}$ well highlights regions likely to contain object instances. However, similarly to the final correction term applied in \cite{Liu_2022_BMVC_countr}, we seek a similarity response that integrates to 1.0 over each object instance, effectively transforming the similarity map into a density map from which the total count can be estimated via integration.

In exemplar-based counting, we can leverage the $N$ exemplar bounding boxes provided as input to properly estimate a normalization factor. In fact, we know that the area occupied by the $N$ exemplars should count up to exactly $N$. 
To this aim, we reuse the elliptical masks $\{\mathbf{M}^{\mathbf{\tilde{b}}_i}\}_{i=1}^N$ introduced in Section~\ref{ssec:feature_extraction} and compute the normalization factor $z$ as follows:
\begin{equation}
\label{eq:norm-factor}
    z = \frac{1}{N} \sum_{(x, y) \in (\bar{L} \times \bar{V})} \mathbf{M}_{xy} \odot \text{minmax}(\mathbf{S})_{xy},
\end{equation}
where the global weighting mask $\mathbf{M}$ is obtained by properly rescaling the masks $\{\mathbf{M}^{\mathbf{\tilde{b}}_i}\}_{i=1}^N$ and accumulating them into a $\bar{L} \times \bar{V}$ map initialized with all zeros.
%
Notice that, to ensure the normalization relies only on positive values, we first apply minmax normalization to $\mathbf{S}$, rescaling it to the range $[0, 1]$. 

The final normalized map is obtained by dividing the minmax-ed activation map by the computed scalar factor $z$:
\begin{equation}
\label{eq:norm}
\hat{\mathbf{S}} = \frac{\text{minmax}(\mathbf{S})}{z}
\end{equation}
This ensures the total response across the $N$ exemplar regions sums to $N$, making the response per object approximate a unit mass. Consequently, the normalized map $\hat{\mathbf{S}}$ serves as a density estimate, with the total count obtained by integrating it over the spatial dimensions.

\subsection{Thresholding density map contributions}
\label{ssec:density_map_thresh}
\newcommand{\patchesArea}{\mathcal{P}\text{atches}\mathcal{A}\text{rea}}

The similarity map highlights object regions but also includes low activations in the background, which can cause overcounting during integration. To address this, we apply a thresholding step to suppress low-activation areas.

In order to find a reasonable threshold, we notice that the normalization criteria detailed in Eqs.~\ref{eq:norm-factor} and~\ref{eq:norm} induce a mean per-patch unit count, estimated on the set $\mathcal{B}$ of exemplars, of $u(\mathcal{B}) = \frac{|\mathcal{B}|}{\sum_{\mathbf{b} \in\mathcal{B}} \text{area}(\mathbf{b})}$ -- where $\text{area}(\cdot)$ is a function returning the area, expressed in the number of patches, for the given bounding box. The resulting term $u(\mathcal{B})$ is the mean unit count value that a patch carries if it is part of a relevant object. This means that it is likely that patches carrying less value than $u(\mathcal{B})$ pertain to a non-relevant object and may be filtered out. 

However, notice that $u(\mathcal{B})$ assumes its minimum value when fed with the set $\mathcal{B^*} = \{\argmax_{\mathbf{b} \in \mathcal{B}} \text{area}(\mathbf{b})\}$ -- i.e. when computed on the set composed solely of the largest exemplar bounding box.
In cases different from this, we are also inevitably filtering out contributions from the other exemplars bounding boxes that we know are valid. For this reason, we set the threshold to be exactly the unit count computed on the largest region, i.e., $\tau = u(\mathcal{B}^*)$.

This process ensures that only the relevant activations are considered, improving the accuracy of the object count estimate.

\subsection{Increasing Spatial Resolution}
\label{ssec:increasing_resolution}
The above-presented method generates consistent results when the spatial resolution of DINO is enough to capture no more than an object per visual patch, which happens when the objects are not too small. Therefore, 
instead of processing the entire image at once, we divide the input image into four non-overlapping and evenly-shaped quadrants and independently process each sub-image with the backbone $\psi_\text{v}$. 
In general, we divide the image into $4^k$ different quadrants, where $k \in \mathbb{N}$ defines the resolution level -- i.e., how many times we recursively partition each quadrant in its four sub-quadrants.  

After extracting features from each quadrant, we spatially reassemble the four feature maps into a single unified feature tensor by stitching them according to their original spatial layout. This yields a higher-resolution feature map $\mathbf{V} \in \mathbb{R}^{2^kL \times 2^kV \times D}$, effectively increasing the granularity of the representation and enabling more precise localization of small objects. 

This modification integrates seamlessly into the pipeline, as 
all subsequent steps -- ROI-Align, similarity computation, and normalization -- remain unchanged.
We found $k=2$ to work well in the presented domain, although we experimented with other values in Section~\ref{ssec:ablation}. 

%% file: sec/4_experiments.tex
\section{Experimental Evaluation}

\subsection{Datasets}
We evaluate our method on two widely-adopted datasets in the class-agnostic counting literature: FSC-147~\cite{DBLP:conf/cvpr/RanjanSNH21} and CARPK~\cite{hsieh2017drone}. 

\paragraph{FSC-147} The FSC-147 dataset~\cite{DBLP:conf/cvpr/RanjanSNH21} contains 6,135 images spanning 147 categories (e.g., plants, animals, vehicles). 
Each image includes three bounding boxes that serve as exemplars for a single object category. The number of objects per image varies significantly, from 7 to 3,731, with an average of 56. The data is split into training (3,659 images, 89 categories), validation (1,286 images, 29 categories), and test sets (1,190 images, 29 categories).

\paragraph{CARPK} The CARPK dataset~\cite{hsieh2017drone} comprises 1,448 drone-view images of parking lots, designed for vehicle counting tasks. It was first employed in the context of CAC by~\cite{DBLP:conf/cvpr/RanjanSNH21}, primarily to evaluate cross-dataset generalization when counting a specific object class -- vehicles. The dataset includes approximately 90,000 annotated vehicles, each labeled with a bounding box. Density maps are generated by placing a point at the center of each bounding box, and a subset of these boxes is used as exemplars. The dataset is split into a training set of 989 images and a test set of 459 images.

\subsection{Metrics}
We evaluate counting performance using two standard metrics: Mean Absolute Error (MAE) and Root Mean Squared Error (RMSE)~\cite{DBLP:conf/eccv/ArtetaLZ16,DBLP:conf/nips/LempitskyZ10,DBLP:conf/cvpr/RanjanSNH21}, defined as
$MAE = \frac{1}{n} \sum_{i=1}^{n} |c_i - \hat{c}_i|$ and
$RMSE = \sqrt{ \frac{1}{n} \sum_{i=1}^{n} (c_i - \hat{c}_i)^2}$,
where $c_i$ and $\hat{c}_i$ are the ground truth and predicted counts for the $i$-th image.



\subsection{Baseline}
Since \ours{} is the first training-free methodology for exemplar-based CAC using an unsupervised backbone, we design a comparison baseline by adapting CutLER~\cite{wang2023cut} -- an unsupervised object detector that leverages pseudo-labels derived from DINO features. As CutLER is not designed for exemplar-based matching, we adapt it to our setting as follows.
We first apply CutLER to the target image to obtain a set of detected object bounding boxes. To filter these detections and retain only those corresponding to the exemplar object category, we extract visual features using DINO -- specifically, DINO ViT-B/8, the same backbone used during CutLER’s training. For each exemplar bounding box, we apply ROI pooling to obtain its feature representation, and average these representations to form a prototype feature vector representing the exemplar class. We then extract features from the detected bounding boxes and compute their similarity to the prototype vector. Detections with a similarity score above a threshold (set to 0.5) are retained, and their count provides the final prediction.




\input{sec/4_table_fsc}

\input{sec/4_table_carpk}

\subsection{Comparison with SOTA}
We quantitatively compare our approach 
against two SOTA training-free CAC approaches -- TFPOC~\cite{DBLP:conf/wacv/Shi0Z24} and OmniCount~\cite{DBLP:journals/corr/abs-2403-05435} -- which, unlike ours, rely on a supervised backbone. We also compare with UnCounTR~\cite{knobel2024uncountr}, which, on the other hand, is label-free but not training-free. UnCounTRv2 is its enhanced version, which employs the DINOv2 backbone and introduces a refinement strategy to improve predictions for images with high object counts.
Finally, we benchmark our method against several fully supervised SOTA CAC methods, including FamNet~\cite{DBLP:conf/cvpr/RanjanSNH21}, RCAC~\cite{DBLP:conf/eccv/GongZ0DS22}, BMNet~\cite{DBLP:conf/cvpr/Shi0FL022}, 
CounTR~\cite{Liu_2022_BMVC_countr},
LOCA~\cite{DBLP:conf/iccv/EukicLZK23}, PseCo~\cite{DBLP:conf/cvpr/HuangD0ZS24}, CACViT~\cite{DBLP:conf/aaai/WangX0024}, and DAVE~\cite{DBLP:conf/cvpr/PelhanLZK24}. For our approach, we implement two variants using different DINO backbones: DINO~\cite{caron2021emerging} ViT-B/8 and DINOv2~\cite{oquab2023dinov2} ViT-L/14 with registers~\cite{darcet2024vision}. 

\begin{figure*}[t]
    \centering
    \includegraphics[width=\textwidth]{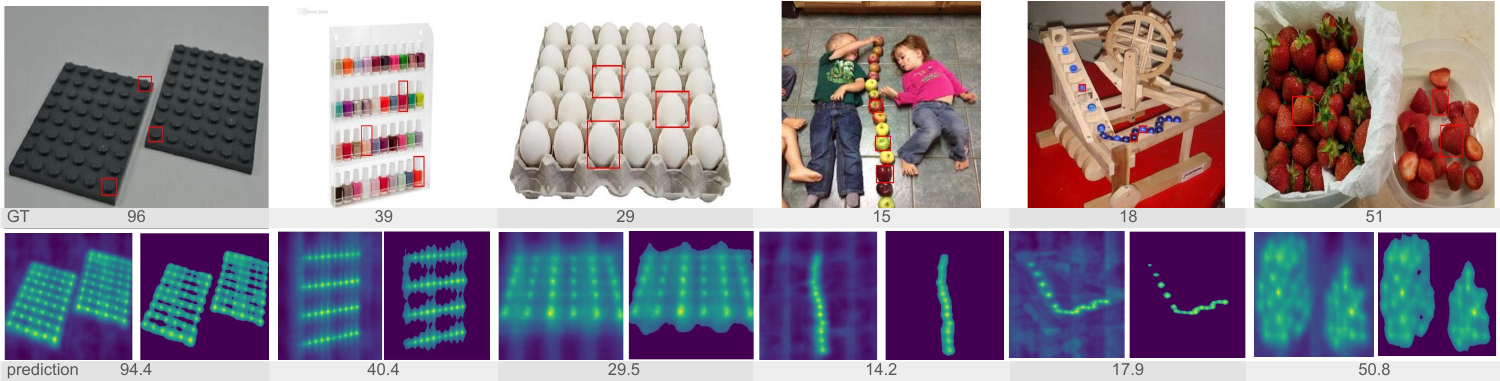}
    \caption{\textbf{Qualitative results on FSC-147.} Samples using DINOv2 ViT L/14 Reg. as backbone. Below the images we report density maps before and after background thresholding.}
    \label{fig:qualitatives}
\end{figure*}

\paragraph{Results on the FSC-147 dataset}
Quantitative results on the FSC-147 validation and test splits are reported in Table~\ref{tab:main_table}, while qualitative examples are shown in Figure~\ref{fig:qualitatives}. Despite being entirely training-free and not relying on human annotations at any stage, our approach -- particularly when using DINOv2 features -- achieves competitive performance. Notably, we outperform the CutLER-based baseline, the only other method operating under the same training- and label-free setting, by a significant margin across all evaluation metrics on both validation and test splits.
Compared to TFPOC and OmniCount -- the two SOTA training-free methods that rely on supervised backbones for feature extraction -- our approach achieves substantially lower RMSE, underscoring its robustness in challenging scenarios involving dense object distributions, occlusions, and overlaps. We also report competitive MAE performance. Furthermore, when compared to UnCounTR  -- which is label-free but not training-free -- our method yields significantly lower errors in both MAE and RMSE across validation and test splits, demonstrating that high-quality counting is achievable without training, by leveraging strong pretrained visual features.
Finally, it is worth emphasizing that our method surpasses several fully supervised CAC methods in terms of RMSE, including FamNet, RCAC, BMNet, and PseCo. This is particularly remarkable given that these methods leverage supervision not only for feature extraction but also for density map regression, relying on costly point-level annotations. These results demonstrate that our approach, despite being training- and label-free, can compete with and even outperform fully supervised alternatives in certain settings.

\paragraph{Results on the CARPK dataset}

Table~\ref{tab:carpk_results} reports the results on CARPK. For FamNet~\cite{DBLP:conf/cvpr/RanjanSNH21}, RCAC~\cite{DBLP:conf/eccv/GongZ0DS22}, BMNet+~\cite{DBLP:conf/cvpr/Shi0FL022}, LOCA~\cite{DBLP:conf/iccv/EukicLZK23}, CACViT~\cite{DBLP:conf/aaai/WangX0024}, and TFPOC~\cite{DBLP:conf/wacv/Shi0Z24}, scores were obtained in the 12-exemplar setting originally proposed by~\cite{DBLP:conf/cvpr/RanjanSNH21}. In contrast, our method, as well as several others, adopts a 3-exemplar setting, which we argue better reflects realistic usage scenarios. Indeed, while providing three visual exemplars of a target class is feasible for most users, collecting twelve is considerably less practical. Moreover, this setting aligns with the protocol used in the FSC-147 dataset.

\ours{} consistently outperforms the CutLER baseline, and with DINOv2 features, it achieves 21.26 MAE and 28.20 RMSE, establishing a new state-of-the-art among unsupervised methods. Notably, our approach also surpasses UnCounTR~\cite{knobel2024uncountr}, despite the latter relying on additional training.

\input{sec/CARPK_qualitatives/qualitativesCARPK_main_0}

Figure~\ref{fig:qualitatives-carpk} presents qualitative results, showing ground-truth counts alongside the density maps predicted by \ours{}. The second row highlights some of the most challenging failure cases. A recurring limitation arises when exemplars include both horizontally and vertically aligned cars: rotated instances tend to elicit weaker activations, leading to systematic undercounting. Similarly, the continuous density maps produced by our model often underestimate counts when objects are partially occluded. Conversely, when an exemplar itself is occluded -- as in the case of the car partially hidden by a bridge in the final example -- the model may overcount, misinterpreting background structures (e.g., the asphalt of the bridge) as part of the object. Another frequent failure occurs when cars are oblique to the camera perspective: exemplar bounding boxes may span multiple instances, resulting in approximately one detection every two cars, as illustrated in the second-to-last example.

\subsection{Ablation Studies}
\label{ssec:ablation}

\paragraph{Effect of the Elliptical Assumption.}
As introduced in Section~\ref{sec:method}, we assume that objects are more likely to appear near the center of exemplar bounding boxes and less likely in the corners. Section \textit{A} of Table~\ref{tab:ablation} shows that incorporating an elliptical weighting mask during feature extraction significantly improves performance. This spatial prior enhances object localization by emphasizing central regions and down-weighting peripheral areas, effectively suppressing background noise. It also benefits the normalization process by concentrating the density around the object and reducing the impact of irrelevant regions.

\paragraph{Effect of the Density Map Thresholding.}
Thresholding the similarity-based density map (Section \ref{ssec:density_map_thresh}) helps suppress spurious background activations. 
Section~\textit{B} of Table~\ref{tab:ablation} shows that this step yields clear improvements in both MAE and RMSE. As shown in Figure~\ref{fig:qualitatives}, it refines density maps by reducing noise and focusing on true object locations.

\newcommand{\original}{Resolution $1\times$}
\newcommand{\highres}{Resolution $2\times$}
\newcommand{\higherres}{Resolution $4\times$}

\begin{table*}[ht]
\centering
\small
\begin{adjustbox}{width=\textwidth}
\begin{tabularx}{\textwidth}{l *{4}{>{\centering\arraybackslash}X}}
    \toprule
     &  \multicolumn{2}{c}{\textbf{Validation}} & \multicolumn{2}{c}{\textbf{Test}} \\
    \cmidrule(r){2-3} \cmidrule(r){4-5}
      & MAE $\downarrow$ & RMSE $\downarrow$ & MAE $\downarrow$ & RMSE $\downarrow$ \\
    \toprule
    \rowcolor{lightgray}
    \multicolumn{5}{l}{\textit{A) Effect of Ellipse assumption}}  \\
    w/o ellipse & 32.32 & 86.83 & 29.73 & 102.38 \\

    \rowcolor{LightCyan}
    w/ ellipse & \textbf{25.48} & \textbf{57.38} & \textbf{20.93} & \textbf{71.37} \\ 
    \midrule
    \rowcolor{lightgray}
    \multicolumn{5}{l}{\textit{B) Effect of thresholding}}  \\
    w/o thresholding & 59.25 & 135.72 & 36.81 & 88.29 \\
    
    \rowcolor{LightCyan}
    w/ thresholding &  \textbf{25.48} & \textbf{57.38} & \textbf{20.93} & \textbf{71.37} \\ 
    
    \midrule
    \rowcolor{lightgray}
    \multicolumn{5}{l}{\textit{C) Effect of changing the resolution}}  \\
    \original & 32.76 & 96.74 & 28.23 & 118.9 \\
    \highres & 25.56 & 66.33 & 23.06 & 90.7 \\

    \rowcolor{LightCyan}
    \higherres &  \textbf{25.48} & \textbf{57.38} & \textbf{20.93} & \textbf{71.37} \\

    \midrule
    \rowcolor{lightgray}
    \multicolumn{5}{l}{\textit{D) Effect of changing the backbone}}  \\
    CLIP ViT B/16 & 139.48 & 383.41 & 72.34 & 196.49 \\
    CLIP ViT L/14 & 155.09 & 414.08 & 80.63 & 205.3 \\
        \hdashline
    MAE ViT B/16 & 121.52 & 327.5 & 57.87 & 171.04 \\
    MAE ViT L/16 & 121.82 & 336.78 & 57.92 & 167.94 \\
        \hdashline
    DINO ResNet50 & 66.18 & 141.87 & 53.87 & 126.46 \\
    DINO ViT S/8 & 115.92 & 280.36 & 55.64 & 158.95 \\
    \rowcolor{LightCyan}
    DINO ViT B/8 & 42.29 & 87.87 & 30.05 & 90.3 \\
        \hdashline
    DINOv2 ViT S/14 & 26.86 & \textbf{57.05} & 22.31 & 79.04 \\
    DINOv2 ViT B/14 & 26.0 & 57.26 & 21.39 & 80.55 \\
    DINOv2 ViT L/14 & \textbf{24.97} & 57.5 & 21.07 & 77.14 \\
    DINOv2 ViT S/14 reg. & 26.68 & 57.29 & \textbf{20.89} & 76.02 \\
    DINOv2 ViT B/14 reg. & 30.45 & 64.79 & 23.84 & 79.7 \\
    \rowcolor{LightCyan}
    DINOv2 ViT L/14 reg. & 25.48 & 57.38 & 20.93 & \textbf{71.37} \\

    \midrule
    \rowcolor{lightgray}
    \multicolumn{5}{l}{\textit{E) Effect of changing the number of exemplars}}  \\
    1 exemplar  & 39.87 & 113.70 & 37.73 & 131.71 \\
    2 exemplars  & 29.18 & 80.68 & 23.12 & 72.80 \\

    \rowcolor{LightCyan}
    3 exemplars & \textbf{25.48} & \textbf{57.38} & \textbf{20.93} & \textbf{71.37} \\ 
    \bottomrule
    \end{tabularx}
\end{adjustbox}
    \caption{\textbf{Ablation study on the FSC-147 dataset.} Impact of various components in our method. 
    Unless specified otherwise, experiments use the DINOv2 ViT L/14 backbone with registers.}

    \label{tab:ablation}
\end{table*}

\paragraph{Effect of Different Resolutions.}
As shown in Section \textit{C} of Table~\ref{tab:ablation}, increasing input resolution via image partitioning consistently improves performance. We evaluate three levels -- $k{=}0$ ($1\times$), $k{=}1$ ($2\times$), and $k{=}2$ ($4\times$). 
Gains are especially notable in RMSE, indicating that higher resolution helps disambiguate small, densely packed objects that may otherwise be lost at coarser scales.

\paragraph{Effect of the visual backbone.}
Beyond DINO and DINOv2, we evaluated alternative backbones, including MAE (unsupervised) and CLIP (trained with web-scale image-text supervision). As shown in Section \textit{D} of Table~\ref{tab:ablation}, performance generally improves with backbone size within each family. DINOv2 consistently outperforms DINO, especially in smaller variants, highlighting its robustness. Registers further boost the DINOv2 model by reducing artifacts that hinder accurate counting. MAE underperforms DINO-based models due to less semantically rich patch features~\cite{barsellotti2024talking}, while CLIP performs worst, confirming prior findings~\cite{barsellotti2024talking, mukhoti2023open} that its features are suboptimal for fine-grained object-centric tasks.


\paragraph{Effect of Varying the Number of Exemplars.}
We assess robustness with fewer exemplars. As shown in Section \textit{E} of Table~\ref{tab:ablation}, increasing the number of exemplars improves counting accuracy, with the largest gain observed from 1 to 2 exemplars. This highlights the benefit of feature diversity and multiple prototypes in capturing appearance variations.



%% file: sec/4_table_fsc.tex
\newcommand{\checkSymbol}{\checkmark}
\newcommand{\xSymbol}{\ding{55}}
\newcommand{\worse}{\textbf{\textcolor{red}{$^{\uparrow}$}}}

\begin{table*}[t]
\centering
    \begin{tabular}{lcccccc}
    \toprule
     &  & & \multicolumn{2}{c}{\textbf{Validation}} & \multicolumn{2}{c}{\textbf{Test}} \\
    \cmidrule(r){4-5} \cmidrule(r){6-7}
    Method & Unsup. & Training-Free & MAE $\downarrow$ & RMSE $\downarrow$ & MAE $\downarrow$ & RMSE $\downarrow$ \\
    \midrule
    FamNet (CVPR '21)~\cite{DBLP:conf/cvpr/RanjanSNH21} & \xSymbol & \xSymbol  & 23.75 & 69.07\worse & 22.08\worse & 99.54\worse\\
    RCAC (ECCV '22)~\cite{DBLP:conf/eccv/GongZ0DS22} & \xSymbol & \xSymbol & 20.54 & 60.78\worse & 20.21 & 81.86\worse\\
    BMNet+ (CVPR '22)~\cite{DBLP:conf/cvpr/Shi0FL022} & \xSymbol & \xSymbol & 15.74 & 58.53\worse & 14.62 & 91.83\worse\\
    CounTR (BMCV '22)~\cite{Liu_2022_BMVC_countr} & \xSymbol & \xSymbol & 13.13 & 49.83 & 11.95 & 91.23\worse \\
    LOCA (ICCV '23)~\cite{DBLP:conf/iccv/EukicLZK23} & \xSymbol & \xSymbol & 10.24 & 32.56 & 10.79 & 56.97 \\
    PseCo (CVPR '24)~\cite{DBLP:conf/cvpr/HuangD0ZS24} & \xSymbol & \xSymbol & 15.31 & 68.34\worse & 13.05 & 112.86\worse\\
    CACViT (AAAI '24)~\cite{DBLP:conf/aaai/WangX0024} & \xSymbol & \xSymbol & 10.63 & 37.95 & 9.13 & 48.96 \\
    DAVE (CVPR '24)~\cite{DBLP:conf/cvpr/PelhanLZK24} & \xSymbol & \xSymbol & \underline{8.91} & \underline{28.08} & \underline{8.66} & \underline{32.36} \\[1pt]
    \hdashline
    \noalign{\vskip 1pt}
    TFPOC (WACV '24)~\cite{DBLP:conf/wacv/Shi0Z24} & \xSymbol & \checkSymbol & - & - & 19.95 & 132.16\worse\\
    OmniCount (AAAI '25)~\cite{DBLP:journals/corr/abs-2403-05435} & \xSymbol & \checkSymbol & - & - & \underline{18.63} & \underline{112.98}\worse \\[1pt]
    \hdashline
    \noalign{\vskip 1pt}
     UnCounTR (CVPR '24)~\cite{knobel2024uncountr} & \checkSymbol & \xSymbol & 36.93\worse & 106.61\worse & 35.77\worse & 130.34\worse \\[1pt]
     UnCounTRv2 (CVPR '24)~\cite{knobel2024uncountr} & \checkSymbol & \xSymbol & - & - & \underline{28.67}\worse & \underline{118.40}\worse \\[1pt]
    \hdashline
    \noalign{\vskip 1pt}
    CutLER Baseline & \checkSymbol & \checkSymbol & 54.18 & 135.29 & 56.44 & 158.01 \\
    \ours \scriptsize (DINO) & \checkSymbol & \checkSymbol & 42.29 & 87.87 & 30.05 & 90.3 \\
    \ours \scriptsize (DINOv2) & \checkSymbol & \checkSymbol & \textbf{25.48} & \textbf{57.38} & \textbf{20.93} & \textbf{71.37} \\
    \bottomrule
    \end{tabular}
    \caption{\textbf{SOTA comparison on FSC-147 (val/test).} Methods are grouped as unsupervised or training-free. Best per category is \underline{underlined}; best training-free method with an unsupervised backbone is in \textbf{bold}; \worse marks results of methods in advantaged (supervised or non-training-free) categories performing worse than \ours{}.}
    \label{tab:main_table}
\end{table*}

%% file: sec/4_table_carpk.tex
\begin{table*}[t]
\centering
\begin{tabular}{lllcc}
\toprule
Method & Unsup. & Training-Free & MAE $\downarrow$ & RMSE $\downarrow$ \\
\midrule
CounTR~\cite{Liu_2022_BMVC_countr} & \xSymbol & \xSymbol & 19.62 & 29.70\worse \\
FamNet (CVPR '21)~\cite{DBLP:conf/cvpr/RanjanSNH21} & \xSymbol & \xSymbol & 28.84\worse &  44.47\worse\\
RCAC (ECCV '22)~\cite{DBLP:conf/eccv/GongZ0DS22} & \xSymbol & \xSymbol & 17.98 & 24.21\\
BMNet+ (CVPR '22)~\cite{DBLP:conf/cvpr/Shi0FL022} & \xSymbol & \xSymbol & 10.44 & 13.77\\
LOCA (ICCV '23)~\cite{DBLP:conf/iccv/EukicLZK23} & \xSymbol & \xSymbol & 9.97 & 12.51 \\
CACViT (AAAI '24)~\cite{DBLP:conf/aaai/WangX0024} & \xSymbol & \xSymbol & \underline{8.30} & \underline{11.18} \\
[1pt]
\midrule
\noalign{\vskip 1pt}
TFPOC (WACV '24)~\cite{DBLP:conf/wacv/Shi0Z24} & \xSymbol & \checkSymbol & \underline{10.97} & \underline{14.24}\\
\midrule
UnCounTR \cite{knobel2024uncountr} & \checkSymbol & \xSymbol & \underline{30.35}\worse & \underline{35.67}\worse \\
\midrule
CutLER Baseline & \checkSymbol & \checkSymbol & 44.06 & 51.31 \\
\ours{} \scriptsize (DINO) & \checkSymbol & \checkSymbol & 36.57 & 47.31 \\ 
\ours{} \scriptsize (DINOv2) & \checkSymbol & \checkSymbol & \textbf{21.26} & \textbf{28.20} \\ 
\bottomrule
\end{tabular}
\caption{\textbf{SOTA comparison on CARPK test split.} Methods are grouped as unsupervised or training-free. Best per category is \underline{underlined}; best training-free method with an unsupervised backbone is in \textbf{bold}; \worse marks results of methods in advantaged (supervised or non-training-free) categories performing worse than \ours{}.}
    \label{tab:carpk_results}
\end{table*}

%% file: sec/CARPK_qualitatives/qualitativesCARPK_main_0.tex
\newcolumntype{C}[1]{>{\centering\arraybackslash}p{#1}}
\arrayrulecolor{gray}

\begin{figure*}[h]
    \centering
    \tiny
    \newcommand{\hd}[1]{\textbf{#1}}
    \newcommand{\figw}{.26\textwidth}
    \setlength{\tabcolsep}{1pt}
    \resizebox{\textwidth}{!}{
    \begin{tabular}{*6{C{\figw}}} 

        \midrule 
        \large
        \figimg{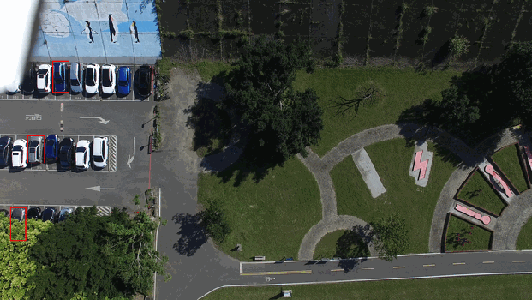} &
        \figimg{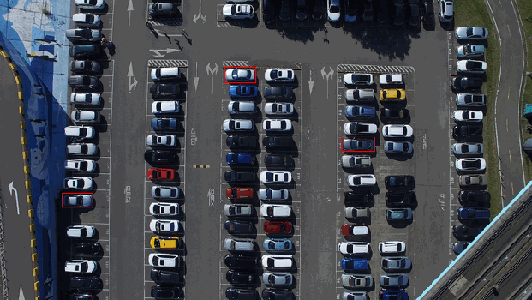} &
        \figimg{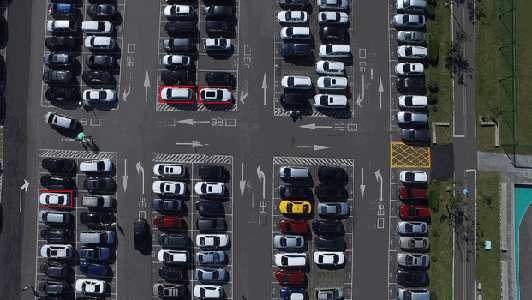} &
        \figimg{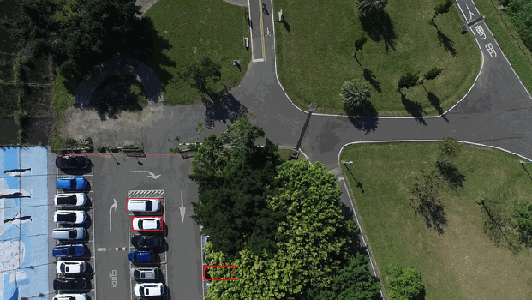} &
        \figimg{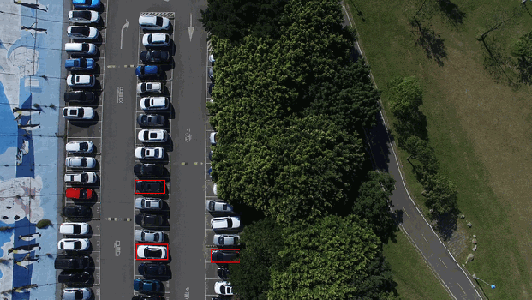} &
        \figimg{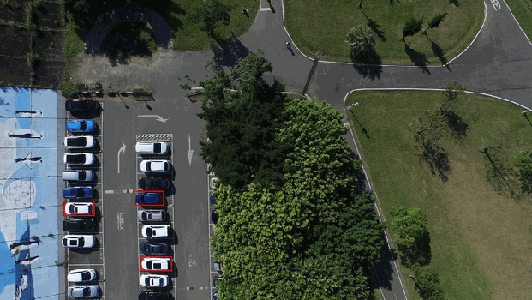} \\
        \rowcolor[gray]{.9}
        21 &
        115 &
        103 &
        19 &
        52 &
        28 \\
        \includegraphics[width=0.45\linewidth, height=1.5cm]{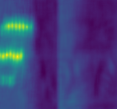}\includegraphics[width=0.45\linewidth, height=1.5cm]{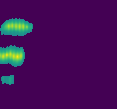} 
 &
        \includegraphics[width=0.45\linewidth, height=1.5cm]{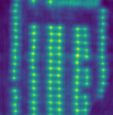}\includegraphics[width=0.45\linewidth, height=1.5cm]{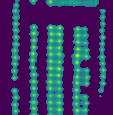} 
 &
        \includegraphics[width=0.45\linewidth, height=1.5cm]{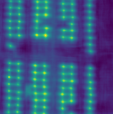}\includegraphics[width=0.45\linewidth, height=1.5cm]{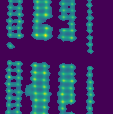} 
 &
        \includegraphics[width=0.45\linewidth, height=1.5cm]{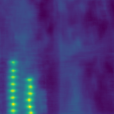}\includegraphics[width=0.45\linewidth, height=1.5cm]{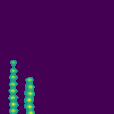} 
 &
        \includegraphics[width=0.45\linewidth, height=1.5cm]{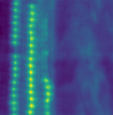}\includegraphics[width=0.45\linewidth, height=1.5cm]{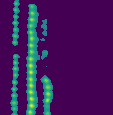} 
 &
        \includegraphics[width=0.45\linewidth, height=1.5cm]{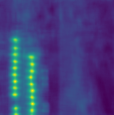}\includegraphics[width=0.45\linewidth, height=1.5cm]{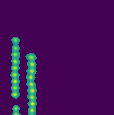} 
 \\
        21.1 &
        115.2 &
        102.1 &
        18.1 &
        50.9 &
        26.8 \\

        \midrule 
        \large
        \figimg{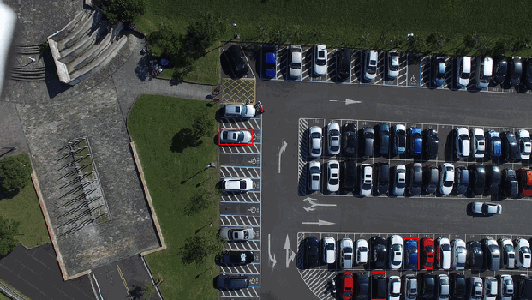} &
        \figimg{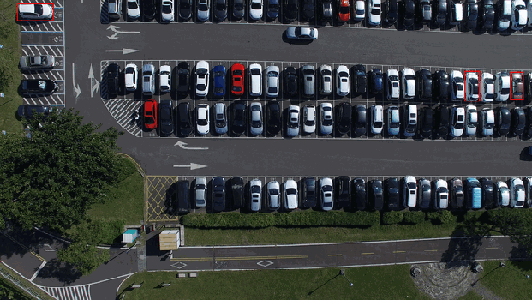} &
        \figimg{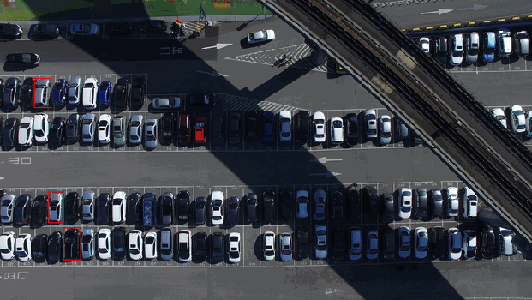} &
        \figimg{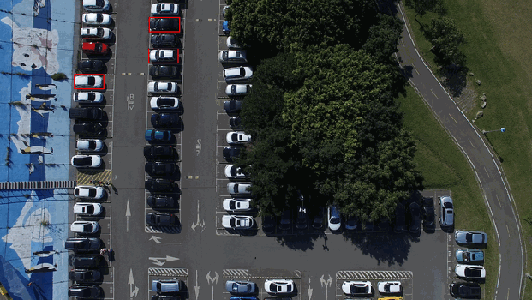} &
        \figimg{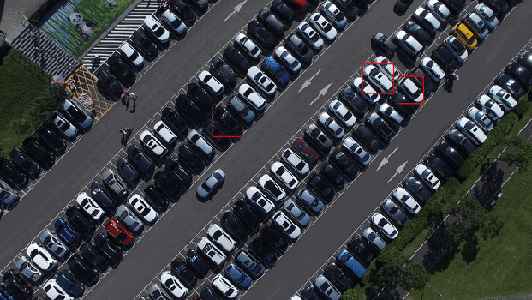} &
        \figimg{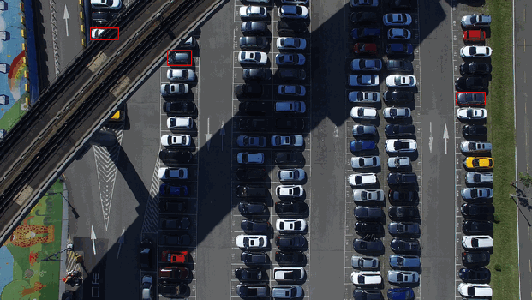} \\
        \rowcolor[gray]{.9}
        73 &
        104 &
        115 &
        73 &
        145 &
        119 \\
        \includegraphics[width=0.45\linewidth, height=1.5cm]{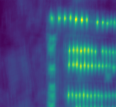}\includegraphics[width=0.45\linewidth, height=1.5cm]{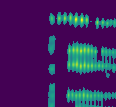} 
 &
        \includegraphics[width=0.45\linewidth, height=1.5cm]{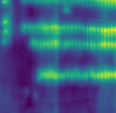}\includegraphics[width=0.45\linewidth, height=1.5cm]{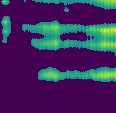} 
 &
        \includegraphics[width=0.45\linewidth, height=1.5cm]{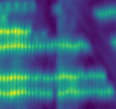}\includegraphics[width=0.45\linewidth, height=1.5cm]{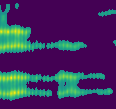} 
 &
        \includegraphics[width=0.45\linewidth, height=1.5cm]{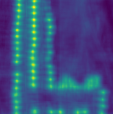}\includegraphics[width=0.45\linewidth, height=1.5cm]{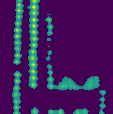} 
 &
        \includegraphics[width=0.45\linewidth, height=1.5cm]{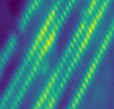}\includegraphics[width=0.45\linewidth, height=1.5cm]{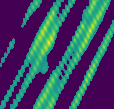} 
 &
        \includegraphics[width=0.45\linewidth, height=1.5cm]{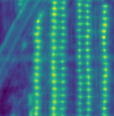}\includegraphics[width=0.45\linewidth, height=1.5cm]{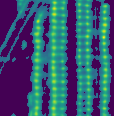} 
 \\
        47.8 &
        64.4 &
        67.0 &
        58.9 &
        66.1 &
        244.0 \\

    \midrule
    \end{tabular}
    }
    \caption{\textbf{Qualitative results on FSC-147.} The last samples were selected according to the highest counting error.}
    \label{fig:qualitatives-carpk}
\end{figure*}

%% file: sec/5_conclusions.tex
\section{Conclusions}
In this paper, we introduced the first \textit{fully unsupervised} and \textit{training-free} exemplar-based framework for class-agnostic counting. By leveraging self-supervised features from DINO, our method addresses a key limitation of existing CAC approaches that depend on costly human annotations and are biased to pay more attention to the labeled categories. 
Our approach achieves competitive performance on the FSC-147 benchmark compared to a baseline under the same setting, training-free methods relying on supervised backbones, unsupervised methods that are not training-free, and even fully supervised approaches.
These results validate the effectiveness of self-supervised representations in counting tasks, paving the way for scalable, annotation- and training-free solutions in open-world scenarios. Future work will refine exemplar quality to further reduce counting noise and extend the method to prompt-based CAC tasks.

%% file: sec/6_ack.tex
\section{Acknowledgements}
This work was partially funded by: Spoke 8, Tuscany Health Ecosystem (THE) Project (CUP B83C22003930001), funded by the National Recovery and Resilience Plan (NRRP), within the NextGeneration Europe (NGEU) Program; Horizon Europe Research \& Innovation Programme under Grant agreement N. 101092612 (Social and hUman ceNtered XR - SUN project); PNRR - M4C2 - Investimento 1.3, Partenariato Esteso PE00000013 - "FAIR - Future Artificial Intelligence Research" - Spoke 1 "Human-centered AI", funded by European Union - NextGenerationEU; ITSERR - ITalian Strengthening of the Esfri Ri Resilience (CUP B53C22001770006), also funded by the European Union via NextGenerationEU.

%% file: sec/X_suppl.tex
\clearpage
\appendix
\setcounter{page}{1}

\section{Supplementary Material}
\addcontentsline{toc}{section}{Supplementary Material}


\section{More Qualitative Results}

In Figures \ref{fig:more-examples-0}, \ref{fig:more-examples-1}, and \ref{fig:more-examples-2}, we report more qualitative results on the FSC147 test split. For each input image, we report i) the image itself, ii) the ground truth total count, iii) the density maps before and after background thresholding, and iv) the counting prediction.

Figures \ref{fig:carpk-more-examples-0}, \ref{fig:carpk-more-examples-1}, \ref{fig:carpk-more-examples-2} and \ref{fig:carpk-more-examples-3} are qualitative results from the CARPK test split with the same formatting.

\input{sec/XX_qualitatives_FSC_0}

\input{sec/XX_qualitatives_FSC_1}

\input{sec/XX_qualitatives_FSC_2}

\newpage
\newpage

\input{sec/CARPK_qualitatives/qualitativesCARPK_0}

\input{sec/CARPK_qualitatives/qualitativesCARPK_1}

\input{sec/CARPK_qualitatives/qualitativesCARPK_2}

\input{sec/CARPK_qualitatives/qualitativesCARPK_3}

%% file: sec/XX_qualitatives_FSC_0.tex
\newcolumntype{C}[1]{>{\centering\arraybackslash}p{#1}}
\arrayrulecolor{gray}

\begin{figure*}[t]
    \centering
    \tiny
    \newcommand{\hd}[1]{\textbf{#1}}
    \newcommand{\figw}{.26\textwidth}
    \setlength{\tabcolsep}{1pt}
    \resizebox{\textwidth}{!}{
\begin{tabular}{*6{C{\figw}}} 

        \midrule 
        \large
        \figimg{images/supplementary/FSC/images/5600.jpg} &
        \figimg{images/supplementary/FSC/images/708.jpg} &
        \figimg{images/supplementary/FSC/images/7173.jpg} &
        \figimg{images/supplementary/FSC/images/5357.jpg} &
        \figimg{images/supplementary/FSC/images/5934.jpg} &
        \figimg{images/supplementary/FSC/images/7295.jpg} \\
        \rowcolor[gray]{.9}
        149 &
        15 &
        48 &
        12 &
        17 &
        63 \\
        \includegraphics[width=0.45\linewidth, height=1.5cm]{images/supplementary/FSC/densitymaps_no_bkg_threshold/5600.jpg}\includegraphics[width=0.45\linewidth, height=1.5cm]{images/supplementary/FSC/densitymaps/5600.jpg} 
 &
        \includegraphics[width=0.45\linewidth, height=1.5cm]{images/supplementary/FSC/densitymaps_no_bkg_threshold/708.jpg}\includegraphics[width=0.45\linewidth, height=1.5cm]{images/supplementary/FSC/densitymaps/708.jpg} 
 &
        \includegraphics[width=0.45\linewidth, height=1.5cm]{images/supplementary/FSC/densitymaps_no_bkg_threshold/7173.jpg}\includegraphics[width=0.45\linewidth, height=1.5cm]{images/supplementary/FSC/densitymaps/7173.jpg} 
 &
        \includegraphics[width=0.45\linewidth, height=1.5cm]{images/supplementary/FSC/densitymaps_no_bkg_threshold/5357.jpg}\includegraphics[width=0.45\linewidth, height=1.5cm]{images/supplementary/FSC/densitymaps/5357.jpg} 
 &
        \includegraphics[width=0.45\linewidth, height=1.5cm]{images/supplementary/FSC/densitymaps_no_bkg_threshold/5934.jpg}\includegraphics[width=0.45\linewidth, height=1.5cm]{images/supplementary/FSC/densitymaps/5934.jpg} 
 &
        \includegraphics[width=0.45\linewidth, height=1.5cm]{images/supplementary/FSC/densitymaps_no_bkg_threshold/7295.jpg}\includegraphics[width=0.45\linewidth, height=1.5cm]{images/supplementary/FSC/densitymaps/7295.jpg} 
 \\
\end{tabular}
    }
    \caption{\textbf{Qualitative results.}}
    \label{fig:more-examples-0}
\end{figure*}

%% file: sec/XX_qualitatives_FSC_1.tex
\newcolumntype{C}[1]{>{\centering\arraybackslash}p{#1}}
\arrayrulecolor{gray}

\begin{figure*}[t]
    \centering
    \tiny
    \newcommand{\hd}[1]{\textbf{#1}}
    \newcommand{\figw}{.26\textwidth}
    \setlength{\tabcolsep}{1pt}
    \resizebox{\textwidth}{!}{
    \begin{tabular}{*6{C{\figw}}} 

        \midrule 
        \large
        \figimg{images/supplementary/FSC/images/5836.jpg} &
        \figimg{images/supplementary/FSC/images/4939.jpg} &
        \figimg{images/supplementary/FSC/images/4073.jpg} &
        \figimg{images/supplementary/FSC/images/7389.jpg} &
        \figimg{images/supplementary/FSC/images/7713.jpg} &
        \figimg{images/supplementary/FSC/images/4916.jpg} \\
        \rowcolor[gray]{.9}
        18 &
        20 &
        25 &
        25 &
        19 &
        91 \\
        \includegraphics[width=0.45\linewidth, height=1.5cm]{images/supplementary/FSC/densitymaps_no_bkg_threshold/5836.jpg}\includegraphics[width=0.45\linewidth, height=1.5cm]{images/supplementary/FSC/densitymaps/5836.jpg} 
 &
        \includegraphics[width=0.45\linewidth, height=1.5cm]{images/supplementary/FSC/densitymaps_no_bkg_threshold/4939.jpg}\includegraphics[width=0.45\linewidth, height=1.5cm]{images/supplementary/FSC/densitymaps/4939.jpg} 
 &
        \includegraphics[width=0.45\linewidth, height=1.5cm]{images/supplementary/FSC/densitymaps_no_bkg_threshold/4073.jpg}\includegraphics[width=0.45\linewidth, height=1.5cm]{images/supplementary/FSC/densitymaps/4073.jpg} 
 &
        \includegraphics[width=0.45\linewidth, height=1.5cm]{images/supplementary/FSC/densitymaps_no_bkg_threshold/7389.jpg}\includegraphics[width=0.45\linewidth, height=1.5cm]{images/supplementary/FSC/densitymaps/7389.jpg} 
 &
        \includegraphics[width=0.45\linewidth, height=1.5cm]{images/supplementary/FSC/densitymaps_no_bkg_threshold/7713.jpg}\includegraphics[width=0.45\linewidth, height=1.5cm]{images/supplementary/FSC/densitymaps/7713.jpg} 
 &
        \includegraphics[width=0.45\linewidth, height=1.5cm]{images/supplementary/FSC/densitymaps_no_bkg_threshold/4916.jpg}\includegraphics[width=0.45\linewidth, height=1.5cm]{images/supplementary/FSC/densitymaps/4916.jpg} 
 \\
\end{tabular}
    }
    \caption{\textbf{Qualitative results.}}
    \label{fig:more-examples-1}
\end{figure*}

%% file: sec/XX_qualitatives_FSC_2.tex
\newcolumntype{C}[1]{>{\centering\arraybackslash}p{#1}}
\arrayrulecolor{gray}

\begin{figure*}[t]
    \centering
    \tiny
    \newcommand{\hd}[1]{\textbf{#1}}
    \newcommand{\figw}{.26\textwidth}
    \setlength{\tabcolsep}{1pt}
    \resizebox{\textwidth}{!}{
    \begin{tabular}{*6{C{\figw}}} 

        \midrule 
        \large
        \figimg{images/supplementary/FSC/images/301.jpg} &
        \figimg{images/supplementary/FSC/images/4079.jpg} &
        \figimg{images/supplementary/FSC/images/5916.jpg} &
        \figimg{images/supplementary/FSC/images/7714.jpg} &
        \figimg{images/supplementary/FSC/images/348.jpg} &
        \figimg{images/supplementary/FSC/images/2910.jpg} \\
        \rowcolor[gray]{.9}
        28 &
        24 &
        38 &
        46 &
        27 &
        14 \\
        \includegraphics[width=0.45\linewidth, height=1.5cm]{images/supplementary/FSC/densitymaps_no_bkg_threshold/301.jpg}\includegraphics[width=0.45\linewidth, height=1.5cm]{images/supplementary/FSC/densitymaps/301.jpg} 
 &
        \includegraphics[width=0.45\linewidth, height=1.5cm]{images/supplementary/FSC/densitymaps_no_bkg_threshold/4079.jpg}\includegraphics[width=0.45\linewidth, height=1.5cm]{images/supplementary/FSC/densitymaps/4079.jpg} 
 &
        \includegraphics[width=0.45\linewidth, height=1.5cm]{images/supplementary/FSC/densitymaps_no_bkg_threshold/5916.jpg}\includegraphics[width=0.45\linewidth, height=1.5cm]{images/supplementary/FSC/densitymaps/5916.jpg} 
 &
        \includegraphics[width=0.45\linewidth, height=1.5cm]{images/supplementary/FSC/densitymaps_no_bkg_threshold/7714.jpg}\includegraphics[width=0.45\linewidth, height=1.5cm]{images/supplementary/FSC/densitymaps/7714.jpg} 
 &
        \includegraphics[width=0.45\linewidth, height=1.5cm]{images/supplementary/FSC/densitymaps_no_bkg_threshold/348.jpg}\includegraphics[width=0.45\linewidth, height=1.5cm]{images/supplementary/FSC/densitymaps/348.jpg} 
 &
        \includegraphics[width=0.45\linewidth, height=1.5cm]{images/supplementary/FSC/densitymaps_no_bkg_threshold/2910.jpg}\includegraphics[width=0.45\linewidth, height=1.5cm]{images/supplementary/FSC/densitymaps/2910.jpg} 
 \\
\end{tabular}
    }
    \caption{\textbf{Qualitative results.}}
    \label{fig:more-examples-2}
\end{figure*}

%% file: sec/CARPK_qualitatives/qualitativesCARPK_0.tex
\newcolumntype{C}[1]{>{\centering\arraybackslash}p{#1}}
\arrayrulecolor{gray}

\begin{figure*}[t]
    \centering
    \tiny
    \newcommand{\hd}[1]{\textbf{#1}}
    \newcommand{\figw}{.26\textwidth}
    \setlength{\tabcolsep}{1pt}
    \resizebox{\textwidth}{!}{
    \begin{tabular}{*6{C{\figw}}} 

        \midrule 
        \large
        \figimg{images/supplementary/CARPK/images/20161225_TPZ_00378.png} &
        \figimg{images/supplementary/CARPK/images/20161225_TPZ_00188.png} &
        \figimg{images/supplementary/CARPK/images/20161225_TPZ_00196.png} &
        \figimg{images/supplementary/CARPK/images/20161225_TPZ_00396.png} &
        \figimg{images/supplementary/CARPK/images/20161225_TPZ_00132.png} &
        \figimg{images/supplementary/CARPK/images/20161225_TPZ_00112.png} \\
        \rowcolor[gray]{.9}
        77 &
        110 &
        17 &
        6 &
        109 &
        73 \\
\end{tabular}
    }
    \caption{\textbf{Qualitative results on CARPK dataset.}}
    \label{fig:carpk-more-examples-0}
\end{figure*}

%% file: sec/CARPK_qualitatives/qualitativesCARPK_1.tex
\newcolumntype{C}[1]{>{\centering\arraybackslash}p{#1}}
\arrayrulecolor{gray}

\begin{figure*}[t]
    \centering
    \tiny
    \newcommand{\hd}[1]{\textbf{#1}}
    \newcommand{\figw}{.26\textwidth}
    \setlength{\tabcolsep}{1pt}
    \resizebox{\textwidth}{!}{
    \begin{tabular}{*6{C{\figw}}} 

        \midrule 
        \large
        \figimg{images/supplementary/CARPK/images/20161225_TPZ_00386.png} &
        \figimg{images/supplementary/CARPK/images/20161225_TPZ_00214.png} &
        \figimg{images/supplementary/CARPK/images/20161225_TPZ_00223.png} &
        \figimg{images/supplementary/CARPK/images/20161225_TPZ_00324.png} &
        \figimg{images/supplementary/CARPK/images/20161225_TPZ_00197.png} &
        \figimg{images/supplementary/CARPK/images/20161225_TPZ_00203.png} \\
        \rowcolor[gray]{.9}
        21 &
        28 &
        53 &
        109 &
        10 &
        6 \\
\end{tabular}
    }
    \caption{\textbf{Qualitative results on CARPK dataset.}}
    \label{fig:carpk-more-examples-1}
\end{figure*}

%% file: sec/CARPK_qualitatives/qualitativesCARPK_2.tex
\newcolumntype{C}[1]{>{\centering\arraybackslash}p{#1}}
\arrayrulecolor{gray}

\begin{figure*}[t]
    \centering
    \tiny
    \newcommand{\hd}[1]{\textbf{#1}}
    \newcommand{\figw}{.26\textwidth}
    \setlength{\tabcolsep}{1pt}
    \resizebox{\textwidth}{!}{
    \begin{tabular}{*6{C{\figw}}} 

        \midrule 
        \large
        \figimg{images/supplementary/CARPK/images/20161225_TPZ_00347.png} &
        \figimg{images/supplementary/CARPK/images/20161225_TPZ_00116.png} &
        \figimg{images/supplementary/CARPK/images/20161225_TPZ_00447.png} &
        \figimg{images/supplementary/CARPK/images/20161225_TPZ_00422.png} &
        \figimg{images/supplementary/CARPK/images/20161225_TPZ_00140.png} &
        \figimg{images/supplementary/CARPK/images/20161225_TPZ_00397.png} \\
        \rowcolor[gray]{.9}
        129 &
        73 &
        104 &
        148 &
        130 &
        13 \\
\end{tabular}
    }
    \caption{\textbf{Qualitative results on CARPK dataset.}}
    \label{fig:carpk-more-examples-2}
\end{figure*}

%% file: sec/CARPK_qualitatives/qualitativesCARPK_3.tex
\newcolumntype{C}[1]{>{\centering\arraybackslash}p{#1}}
\arrayrulecolor{gray}

\begin{figure*}[t]
    \centering
    \tiny
    \newcommand{\hd}[1]{\textbf{#1}}
    \newcommand{\figw}{.26\textwidth}
    \setlength{\tabcolsep}{1pt}
    \resizebox{\textwidth}{!}{
    \begin{tabular}{*6{C{\figw}}} 

        \midrule 
        \large
        \figimg{images/supplementary/CARPK/images/20161225_TPZ_00204.png} &
        \figimg{images/supplementary/CARPK/images/20161225_TPZ_00198.png} &
        \figimg{images/supplementary/CARPK/images/20161225_TPZ_00392.png} &
        \figimg{images/supplementary/CARPK/images/20161225_TPZ_00310.png} &
        \figimg{images/supplementary/CARPK/images/20161225_TPZ_00199.png} &
        \figimg{images/supplementary/CARPK/images/20161225_TPZ_00006.png} \\
        \rowcolor[gray]{.9}
        7 &
        4 &
        2 &
        75 &
        4 &
        86 \\
\end{tabular}
    }
    \caption{\textbf{Qualitative results on CARPK dataset.}}
    \label{fig:carpk-more-examples-3}
\end{figure*}

%% file: main.bbl
\begin{thebibliography}{36}
\providecommand{\natexlab}[1]{#1}
\providecommand{\url}[1]{\texttt{#1}}
\expandafter\ifx\csname urlstyle\endcsname\relax
  \providecommand{\doi}[1]{doi: #1}\else
  \providecommand{\doi}{doi: \begingroup \urlstyle{rm}\Url}\fi

\bibitem[Amato et~al.(2019)Amato, Ciampi, Falchi, and Gennaro]{DBLP:conf/iscc/AmatoCFG19}
Giuseppe Amato, Luca Ciampi, Fabrizio Falchi, and Claudio Gennaro.
\newblock Counting vehicles with deep learning in onboard {UAV} imagery.
\newblock In \emph{2019 {IEEE} Symposium on Computers and Communications, {ISCC} 2019, Barcelona, Spain, June 29 - July 3, 2019}, pages 1--6. {IEEE}, 2019.

\bibitem[Arteta et~al.(2016)Arteta, Lempitsky, and Zisserman]{DBLP:conf/eccv/ArtetaLZ16}
Carlos Arteta, Victor~S. Lempitsky, and Andrew Zisserman.
\newblock Counting in the wild.
\newblock In \emph{ECCV}, pages 483--498. Springer, 2016.

\bibitem[Barsellotti et~al.(2024)Barsellotti, Bianchi, Messina, Carrara, Cornia, Baraldi, Falchi, and Cucchiara]{barsellotti2024talking}
Luca Barsellotti, Lorenzo Bianchi, Nicola Messina, Fabio Carrara, Marcella Cornia, Lorenzo Baraldi, Fabrizio Falchi, and Rita Cucchiara.
\newblock Talking to dino: Bridging self-supervised vision backbones with language for open-vocabulary segmentation.
\newblock \emph{arXiv preprint arXiv:2411.19331}, 2024.

\bibitem[Benedetto et~al.(2022)Benedetto, Carrara, Ciampi, Falchi, Gennaro, and Amato]{DBLP:journals/eswa/BenedettoCCFGA22}
Marco~Di Benedetto, Fabio Carrara, Luca Ciampi, Fabrizio Falchi, Claudio Gennaro, and Giuseppe Amato.
\newblock An embedded toolset for human activity monitoring in critical environments.
\newblock \emph{Expert Syst. Appl.}, 199:\penalty0 117125, 2022.

\bibitem[Bereciartua{-}Perez et~al.(2022)Bereciartua{-}Perez, G{\'{o}}mez, Pic{\'{o}}n, Navarra{-}Mestre, Klukas, and Eggers]{DBLP:journals/cea/Bereciartua-Perez22}
Arantza Bereciartua{-}Perez, Laura G{\'{o}}mez, Artzai Pic{\'{o}}n, Ram{\'{o}}n Navarra{-}Mestre, Christian Klukas, and Till Eggers.
\newblock Insect counting through deep learning-based density maps estimation.
\newblock \emph{Comput. Electron. Agric.}, 197:\penalty0 106933, 2022.

\bibitem[Caron et~al.(2021)Caron, Touvron, Misra, J{\'e}gou, Mairal, Bojanowski, and Joulin]{caron2021emerging}
Mathilde Caron, Hugo Touvron, Ishan Misra, Herv{\'e} J{\'e}gou, Julien Mairal, Piotr Bojanowski, and Armand Joulin.
\newblock {Emerging Properties in Self-Supervised Vision Transformers}.
\newblock In \emph{ICCV}, 2021.

\bibitem[Ciampi et~al.(2022)Ciampi, Carrara, Totaro, Mazziotti, Lupori, Santiago, Amato, Pizzorusso, and Gennaro]{DBLP:journals/mia/CiampiCTMLSAPG22}
Luca Ciampi, Fabio Carrara, Valentino Totaro, Raffaele Mazziotti, Leonardo Lupori, Carlos Santiago, Giuseppe Amato, Tommaso Pizzorusso, and Claudio Gennaro.
\newblock Learning to count biological structures with raters' uncertainty.
\newblock \emph{Medical Image Anal.}, 80:\penalty0 102500, 2022.

\bibitem[Ciampi et~al.(2023)Ciampi, Zeni, Incrocci, Canale, Benelli, Falchi, Amato, and Chessa]{DBLP:journals/ecoi/CiampiZICBFAC23}
Luca Ciampi, Valeria Zeni, Luca Incrocci, Angelo Canale, Giovanni Benelli, Fabrizio Falchi, Giuseppe Amato, and Stefano Chessa.
\newblock A deep learning-based pipeline for whitefly pest abundance estimation on chromotropic sticky traps.
\newblock \emph{Ecol. Informatics}, 78:\penalty0 102384, 2023.

\bibitem[Ciampi et~al.(2025{\natexlab{a}})Ciampi, Azmoudeh, Akbaba, Saritas, Yazici, Ekenel, Amato, and Falchi]{DBLP:journals/corr/abs-2501-19184}
Luca Ciampi, Ali Azmoudeh, Elif~Ecem Akbaba, Erdi Saritas, Ziya~Ata Yazici, Hazim~Kemal Ekenel, Giuseppe Amato, and Fabrizio Falchi.
\newblock A survey on class-agnostic counting: Advancements from reference-based to open-world text-guided approaches.
\newblock \emph{CoRR}, abs/2501.19184, 2025{\natexlab{a}}.

\bibitem[Ciampi et~al.(2025{\natexlab{b}})Ciampi, Messina, Pierucci, Amato, Avvenuti, and Falchi]{10944113}
Luca Ciampi, Nicola Messina, Matteo Pierucci, Giuseppe Amato, Marco Avvenuti, and Fabrizio Falchi.
\newblock Mind the prompt: A novel benchmark for prompt-based class-agnostic counting.
\newblock In \emph{WACV}, pages 7970--7979, 2025{\natexlab{b}}.

\bibitem[Darcet et~al.(2024)Darcet, Oquab, Mairal, and Bojanowski]{darcet2024vision}
Timoth{\'e}e Darcet, Maxime Oquab, Julien Mairal, and Piotr Bojanowski.
\newblock {Vision Transformers Need Registers}.
\newblock In \emph{ICLR}, 2024.

\bibitem[Dhukic et~al.(2023)Dhukic, Lukezic, Zavrtanik, and Kristan]{DBLP:conf/iccv/EukicLZK23}
Nikola Dhukic, Alan Lukezic, Vitjan Zavrtanik, and Matej Kristan.
\newblock A low-shot object counting network with iterative prototype adaptation.
\newblock In \emph{ICCV}, pages 18826--18835. {IEEE}, 2023.

\bibitem[Dosovitskiy et~al.(2021)Dosovitskiy, Beyer, Kolesnikov, Weissenborn, Zhai, Unterthiner, Dehghani, Minderer, Heigold, Gelly, Uszkoreit, and Houlsby]{dosovitskiy2021an}
Alexey Dosovitskiy, Lucas Beyer, Alexander Kolesnikov, Dirk Weissenborn, Xiaohua Zhai, Thomas Unterthiner, Mostafa Dehghani, Matthias Minderer, Georg Heigold, Sylvain Gelly, Jakob Uszkoreit, and Neil Houlsby.
\newblock {An Image is Worth 16x16 Words: Transformers for Image Recognition at Scale}.
\newblock In \emph{ICLR}, 2021.

\bibitem[Gong et~al.(2022)Gong, Zhang, Yang, Dai, and Schiele]{DBLP:conf/eccv/GongZ0DS22}
Shenjian Gong, Shanshan Zhang, Jian Yang, Dengxin Dai, and Bernt Schiele.
\newblock Class-agnostic object counting robust to intraclass diversity.
\newblock In \emph{ECCV}, pages 388--403. Springer, 2022.

\bibitem[He et~al.(2016)He, Zhang, Ren, and Sun]{DBLP:conf/cvpr/HeZRS16}
Kaiming He, Xiangyu Zhang, Shaoqing Ren, and Jian Sun.
\newblock Deep residual learning for image recognition.
\newblock In \emph{CVPR}, pages 770--778. {IEEE}, 2016.

\bibitem[He et~al.(2020)He, Gkioxari, Dollár, and Girshick]{8372616}
Kaiming He, Georgia Gkioxari, Piotr Dollár, and Ross Girshick.
\newblock Mask r-cnn.
\newblock \emph{IEEE TPAMI}, 42\penalty0 (2):\penalty0 386--397, 2020.

\bibitem[He et~al.(2022)He, Chen, Xie, Li, Doll{\'a}r, and Girshick]{he2022masked}
Kaiming He, Xinlei Chen, Saining Xie, Yanghao Li, Piotr Doll{\'a}r, and Ross Girshick.
\newblock Masked autoencoders are scalable vision learners.
\newblock In \emph{CVPR}, pages 16000--16009, 2022.

\bibitem[Hsieh et~al.(2017)Hsieh, Lin, and Hsu]{hsieh2017drone}
Meng-Ru Hsieh, Yen-Liang Lin, and Winston~H Hsu.
\newblock Drone-based object counting by spatially regularized regional proposal network.
\newblock In \emph{Proceedings of the IEEE international conference on computer vision}, pages 4145--4153, 2017.

\bibitem[Huang et~al.(2024)Huang, Dai, Zhang, Zhang, and Shan]{DBLP:conf/cvpr/HuangD0ZS24}
Zhizhong Huang, Mingliang Dai, Yi Zhang, Junping Zhang, and Hongming Shan.
\newblock Point, segment and count: {A} generalized framework for object counting.
\newblock In \emph{CVPR}, pages 17067--17076. {IEEE}, 2024.

\bibitem[Kirillov et~al.(2023)Kirillov, Mintun, Ravi, Mao, Rolland, Gustafson, Xiao, Whitehead, Berg, Lo, Doll{\'{a}}r, and Girshick]{DBLP:conf/iccv/KirillovMRMRGXW23}
Alexander Kirillov, Eric Mintun, Nikhila Ravi, Hanzi Mao, Chlo{\'{e}} Rolland, Laura Gustafson, Tete Xiao, Spencer Whitehead, Alexander~C. Berg, Wan{-}Yen Lo, Piotr Doll{\'{a}}r, and Ross~B. Girshick.
\newblock Segment anything.
\newblock In \emph{ICCV}, pages 3992--4003. {IEEE}, 2023.

\bibitem[Knobel et~al.(2024)Knobel, Han, and Asano]{knobel2024uncountr}
Lukas Knobel, Tengda Han, and Yuki~M. Asano.
\newblock Learning to count without annotations.
\newblock In \emph{{IEEE/CVF} Conference on Computer Vision and Pattern Recognition, {CVPR} 2024, Seattle, WA, USA, June 16-22, 2024}, pages 22924--22934. {IEEE}, 2024.

\bibitem[Lempitsky and Zisserman(2010)]{DBLP:conf/nips/LempitskyZ10}
Victor~S. Lempitsky and Andrew Zisserman.
\newblock Learning to count objects in images.
\newblock In \emph{NeurIPS}, pages 1324--1332. Curran Associates, Inc., 2010.

\bibitem[Liu et~al.(2022)Liu, Zhong, Zisserman, and Xie]{Liu_2022_BMVC_countr}
Chang Liu, Yujie Zhong, Andrew Zisserman, and Weidi Xie.
\newblock Countr: Transformer-based generalised visual counting.
\newblock In \emph{33rd British Machine Vision Conference 2022, {BMVC} 2022, London, UK, November 21-24, 2022}. {BMVA} Press, 2022.

\bibitem[Liu et~al.(2019)Liu, Salzmann, and Fua]{DBLP:conf/cvpr/LiuSF19}
Weizhe Liu, Mathieu Salzmann, and Pascal Fua.
\newblock Context-aware crowd counting.
\newblock In \emph{CVPR}, pages 5099--5108. Computer Vision Foundation / {IEEE}, 2019.

\bibitem[Mondal et~al.(2025 (Accepted - To appear))Mondal, Nag, Zhu, and Dutta]{DBLP:journals/corr/abs-2403-05435}
Anindya Mondal, Sauradip Nag, Xiatian Zhu, and Anjan Dutta.
\newblock Omnicount: Multi-label object counting with semantic-geometric priors.
\newblock 2025 (Accepted - To appear).

\bibitem[Mukhoti et~al.(2023)Mukhoti, Lin, Poursaeed, Wang, Shah, Torr, and Lim]{mukhoti2023open}
Jishnu Mukhoti, Tsung-Yu Lin, Omid Poursaeed, Rui Wang, Ashish Shah, Philip~HS Torr, and Ser-Nam Lim.
\newblock Open vocabulary semantic segmentation with patch aligned contrastive learning.
\newblock In \emph{CVPR}, pages 19413--19423, 2023.

\bibitem[Oquab et~al.(2023)Oquab, Darcet, Moutakanni, Vo, Szafraniec, Khalidov, Fernandez, Haziza, Massa, El-Nouby, et~al.]{oquab2023dinov2}
Maxime Oquab, Timoth{\'e}e Darcet, Th{\'e}o Moutakanni, Huy Vo, Marc Szafraniec, Vasil Khalidov, Pierre Fernandez, Daniel Haziza, Francisco Massa, Alaaeldin El-Nouby, et~al.
\newblock {DINOv2: Learning Robust Visual Features without Supervision}.
\newblock \emph{arXiv preprint arXiv:2304.07193}, 2023.

\bibitem[Pelhan et~al.(2024)Pelhan, Lukezic, Zavrtanik, and Kristan]{DBLP:conf/cvpr/PelhanLZK24}
Jer Pelhan, Alan Lukezic, Vitjan Zavrtanik, and Matej Kristan.
\newblock {DAVE} - {A} detect-and-verify paradigm for low-shot counting.
\newblock In \emph{CVPR}, pages 23293--23302. {IEEE}, 2024.

\bibitem[Ranjan et~al.(2021)Ranjan, Sharma, Nguyen, and Hoai]{DBLP:conf/cvpr/RanjanSNH21}
Viresh Ranjan, Udbhav Sharma, Thu Nguyen, and Minh Hoai.
\newblock Learning to count everything.
\newblock In \emph{CVPR}, pages 3394--3403. Computer Vision Foundation / {IEEE}, 2021.

\bibitem[Shi et~al.(2022)Shi, Lu, Feng, Liu, and Cao]{DBLP:conf/cvpr/Shi0FL022}
Min Shi, Hao Lu, Chen Feng, Chengxin Liu, and Zhiguo Cao.
\newblock Represent, compare, and learn: {A} similarity-aware framework for class-agnostic counting.
\newblock In \emph{CVPR}, pages 9519--9528. {IEEE}, 2022.

\bibitem[Shi et~al.(2024)Shi, Sun, and Zhang]{DBLP:conf/wacv/Shi0Z24}
Zenglin Shi, Ying Sun, and Mengmi Zhang.
\newblock Training-free object counting with prompts.
\newblock In \emph{WACV}, pages 322--330. {IEEE}, 2024.

\bibitem[Tian et~al.(2019)Tian, Guo, Chen, Wang, Long, and Ma]{DBLP:journals/cea/TianGCWLM19}
Mengxiao Tian, Hao Guo, Hong Chen, Qing Wang, Chengjiang Long, and Yuhao Ma.
\newblock Automated pig counting using deep learning.
\newblock \emph{Comput. Electron. Agric.}, 163, 2019.

\bibitem[Wang et~al.(2023)Wang, Girdhar, Yu, and Misra]{wang2023cut}
Xudong Wang, Rohit Girdhar, Stella~X Yu, and Ishan Misra.
\newblock Cut and learn for unsupervised object detection and instance segmentation.
\newblock In \emph{CVPR}, pages 3124--3134, 2023.

\bibitem[Wang et~al.(2024)Wang, Xiao, Cao, and Lu]{DBLP:conf/aaai/WangX0024}
Zhicheng Wang, Liwen Xiao, Zhiguo Cao, and Hao Lu.
\newblock Vision transformer off-the-shelf: {A} surprising baseline for few-shot class-agnostic counting.
\newblock In \emph{AAAI}, pages 5832--5840. {AAAI} Press, 2024.

\bibitem[Xue et~al.(2016)Xue, Ray, Hugh, and Bigras]{DBLP:conf/eccv/XueRHB16}
Yao Xue, Nilanjan Ray, Judith Hugh, and Gilbert Bigras.
\newblock Cell counting by regression using convolutional neural network.
\newblock In \emph{ECCV}, pages 274--290. Springer, 2016.

\bibitem[Zhang et~al.(2017)Zhang, Wu, Costeira, and Moura]{DBLP:conf/iccv/ZhangWCM17}
Shanghang Zhang, Guanhang Wu, Jo{\~{a}}o~Paulo Costeira, and Jos{\'{e}} M.~F. Moura.
\newblock Fcn-rlstm: Deep spatio-temporal neural networks for vehicle counting in city cameras.
\newblock In \emph{ICCV}, pages 3687--3696. {IEEE} Computer Society, 2017.

\end{thebibliography}
